%% file: egpaper_final.tex
\documentclass[10pt,twocolumn,letterpaper]{article}

\usepackage{iccv}
\usepackage{times}
\usepackage{epsfig}
\usepackage{graphicx}
\usepackage{amsmath}
\usepackage{amssymb}
\usepackage{booktabs}
\usepackage{multirow}
\usepackage{placeins}
\usepackage[breaklinks=true,bookmarks=false]{hyperref}
\usepackage[title]{appendix}
\iccvfinalcopy

\def\iccvPaperID{2300}

\newcommand{\set}[1]{\ensuremath{\mathcal{#1}}}
\newcommand{\con}[1]{#1} 

\newcommand{\argmax}{\operatornamewithlimits{\arg\,\max}}

\def\eg{\textit{e.g.}~}
\def\ie{\textit{i.e.}~}

\def\etal{\textit{et al.}~}

\newcommand{\DAL}{DA-layer\xspace }
\newcommand{\DALs}{DA-layers\xspace }
\newcommand{\DIAL}{AutoDIAL\xspace }
\newcommand{\Alex}{AlexNet\xspace }
\newcommand{\Inception}{Inception-BN\xspace }

\newcommand{\DIALAlex}{\DIAL{} -- \Alex{}}
\newcommand{\DIALInception}{\DIAL{} -- \Inception{}}

\begin{document}

\title{AutoDIAL: Automatic DomaIn Alignment Layers}

\author{Fabio Maria Carlucci\\
Sapienza, Roma, Italy
\and
Lorenzo Porzi\\
IRI CSIC-UPC, Barcelona, Spain\\
Mapillary, Graz, Austria
\and
Barbara Caputo\\
Sapienza, Roma, Italy
\and
Elisa Ricci\\
FBK, Trento, Italy\\
University of Perugia, Italy
\and
Samuel Rota Bul\`o\\
FBK, Trento, Italy\\
Mapillary, Graz, Austria
}

\maketitle
\thispagestyle{empty}

\begin{abstract}
Classifiers trained on given databases perform poorly when tested on data acquired in different settings. This is explained in domain adaptation through a shift among distributions of the source and target domains.
Attempts to align them have traditionally resulted in works reducing the domain shift by introducing appropriate loss terms, measuring the discrepancies between source and target distributions, in the objective function.
Here we take a different route, proposing to align the learned representations by embedding in any given network specific \emph{Domain Alignment Layers}, designed to match the source and target feature distributions to a reference one.
Opposite to previous works which define \emph{a priori} in which layers adaptation should be performed, our method is able to \emph{automatically} learn the degree of feature alignment required at different levels of the deep network.
Thorough experiments on different public benchmarks, in the unsupervised setting, confirm the power of our approach.
\end{abstract}

\section{Introduction}
\label{intro.tex}
\input{intro}
\section{Related Work}
\label{related}
\input{related.tex}

\section{Automatic DomaIn Alignment Layers}
\input{method.tex}

\section{Experiments}
\label{experiments}

In this section we extensively evaluate our approach and compare it with state of the art unsupervised domain adaptation methods.
We also provide a detailed analysis of the proposed framework, demonstrating empirically the effect of our contributions.
Note that all the results in the following are reported as averages over five training/testing runs.

\subsection{Experimental Setup}
\label{sec:setup}
\paragraph{Datasets.} We evaluate the proposed approach on three publicly-available datasets. 

The \textbf{Office 31}\cite{saenko2010adapting} dataset is a standard benchmark for testing domain-adaptation methods. It contains 4652 images organized in 31 classes from three different domains: Amazon (A), DSRL (D) and Webcam (W). Amazon images are collected from \texttt{amazon.com}, Webcam and DSLR images were manually gathered in an office environment. In our experiments we consider all possible source\slash{}target combinations of these domains and adopt the \emph{full protocol} setting \cite{gong2013connecting}, \ie we train on the entire labeled source and unlabeled target data and test on annotated target samples.

The \textbf{Office-Caltech} \cite{gong2012geodesic} dataset is obtained by selecting the subset of $10$ common categories in the Office31 and the Caltech256\cite{griffin2007caltech} datasets.
It contains $2533$ images of which about half belong to Caltech256.
Each of Amazon (A), DSLR (D), Webcam (W) and Caltech256 (C) are regarded as separate domains.
In our experiments we only consider the source\slash{}target combinations containing C as either the source or target domain.

To further perform an analysis on a large-scale dataset, we also consider the recent \textbf{Cross-Dataset Testbed} introduced in \cite{tommasi2014testbed} and specifically the \textbf{Caltech-ImageNet} setting. This dataset was obtained by collecting the images corresponding to the $40$ classes shared between the Caltech256 (C) and the Imagenet (I) \cite{deng2009imagenet} datasets. To facilitate comparison with previous works \cite{tzeng2015simultaneous,tommasi2016learning,sun2016return} we perform experiments in two different settings. The first setting, adopted in \cite{tommasi2016learning,tzeng2015simultaneous}, considers 5 splits obtained by selecting 5534 images from ImageNet and 4366 images from Caltech256 across all 40 categories. The second setting, adopted in \cite{sun2016return}, uses 3847 images for Caltech256 and 4000 images for ImageNet.

\paragraph{Networks and Training.}
We apply the proposed method to two state of the art CNNs, \ie AlexNet~\cite{krizhevsky2012imagenet} and Inception-BN~\cite{ioffe2015batch}.
We train our networks using mini-batch stochastic gradient descent with momentum, as implemented in the Caffe library, using the following meta-parameters: weight decay $5\times 10^{-4}$, momentum $0.9$, initial learning rate $10^{-3}$.
We augment the input data by scaling all images to $256\times 256$ pxls, randomly cropping $227\times 227$ pxls (for AlexNet) or $224\times 224$ pxls (Inception-BN) patches and performing random flips.
In all experiments we choose the parameter $\lambda$ by cross-validation on the source set according to the protocol in~\cite{long2016unsupervised}. 

AlexNet~\cite{krizhevsky2012imagenet} is a well-know architecture with five convolutional and three fully-connected layers, denoted as \texttt{fc6}, \texttt{fc7} and \texttt{fc8}.
The outputs of \texttt{fc6} and \texttt{fc7} are commonly used in the domain-adaptation literature as pre-trained feature representations~\cite{donahue2014decaf,sun2016return} for traditional machine learning approaches.
In our experiments we modify AlexNet by appending a \DAL to each fully-connected layer.
Differently from the original AlexNet, we \emph{do not} perform dropout on the outputs of \texttt{fc6} and \texttt{fc7}.
We initialize the network parameters from a publicly-available model trained on the ILSVRC-2012 data, we freeze all the convolutional layers, and increase the learning rate of \texttt{fc8} by a factor of $10$.
During training, each mini-batch contains a number of source and target samples proportional to the size of the corresponding dataset, while the batch size remains fixed at 256.
We train for a total of 60 epochs (where ``epoch'' refers to a complete pass over the source set), reducing the learning rate by a factor $10$ after 54 epochs.

Inception-BN~\cite{ioffe2015batch} is a very deep architecture obtained by concatenating ``inception'' blocks.
Each block is composed of several parallel convolutions with batch normalization and pooling layers.
To apply the proposed method to Inception-BN, we replace each batch-normalization layer with a \DAL.
Similarly to AlexNet, we initialize the network's parameters from a publicly-available model trained on the ILSVRC-2012 data and freeze the first three inception blocks.
The $\alpha$ parameter is also fixed to a value of $0.5$ in the \DALs of the first three blocks, which is equivalent to preserving the original batch normalization layers.
Due to GPU memory constraints, we use a much smaller batch size than for AlexNet and fix the number of source and target samples in each batch to, respectively, 32 and 16.
In the Office-31 experiments we train for 1200 iterations, reducing the learning rate by a factor 10 after 1000 iterations, while in the Cross-Dataset Testbed experiments we train for 2000 iterations, reducing the learning rate after 1500.

\subsection{Analysis of the proposed method}
\label{sec:results-analysis}

\begin{figure}
  \centering
  \includegraphics[width=0.9\columnwidth]{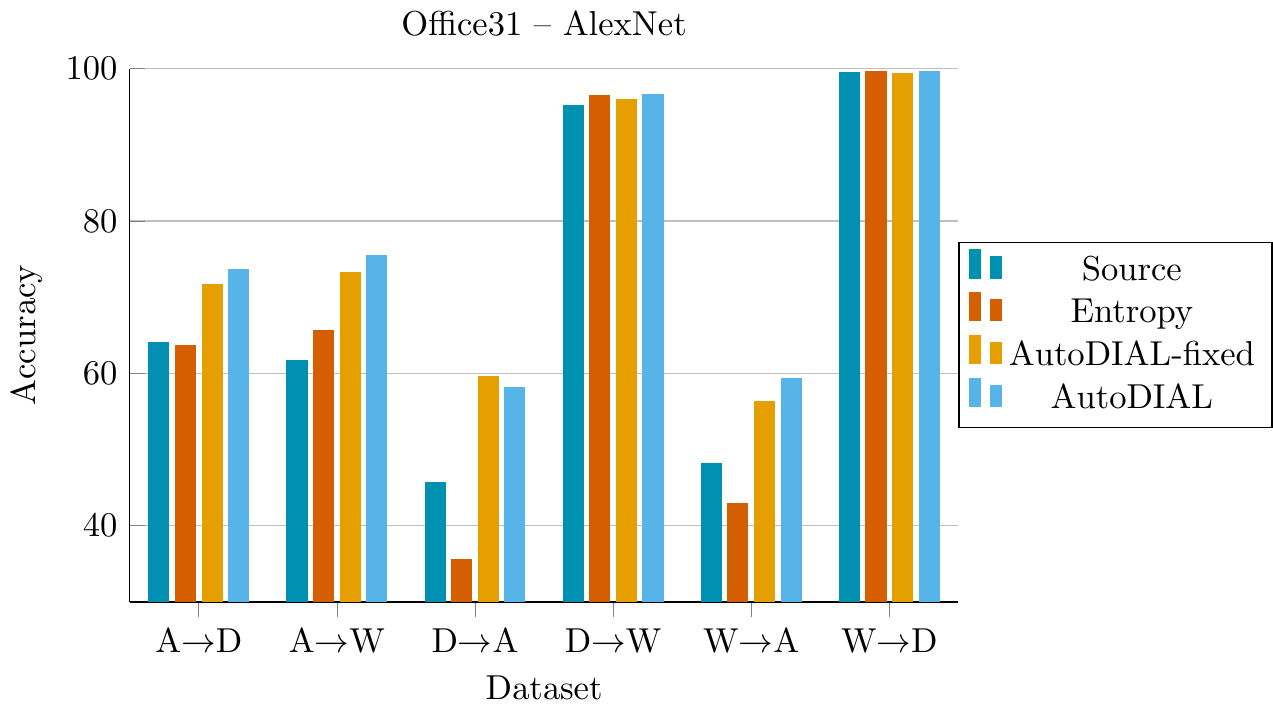}
  \caption{Accuracy on the Office31 dataset when considering different architectures based on AlexNet. 
  }\vspace{-10pt}
  \label{fig:ablation}
\end{figure}

\begin{figure*}
  \centering
  \includegraphics[height=3.5cm]{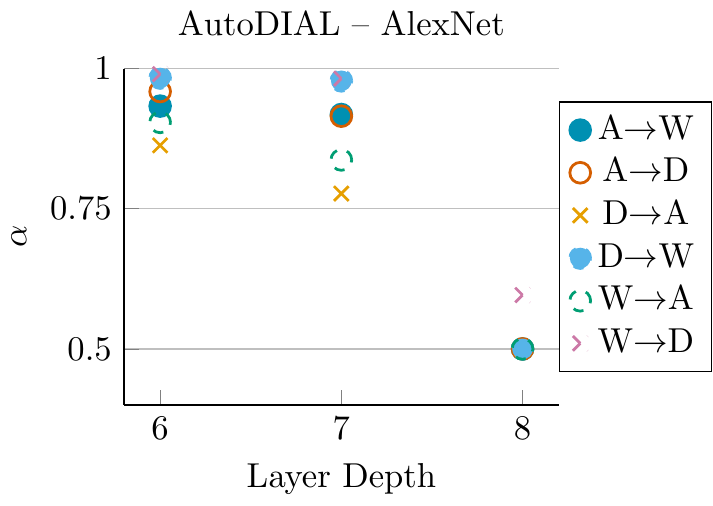}\hfill
  \includegraphics[height=3.5cm]{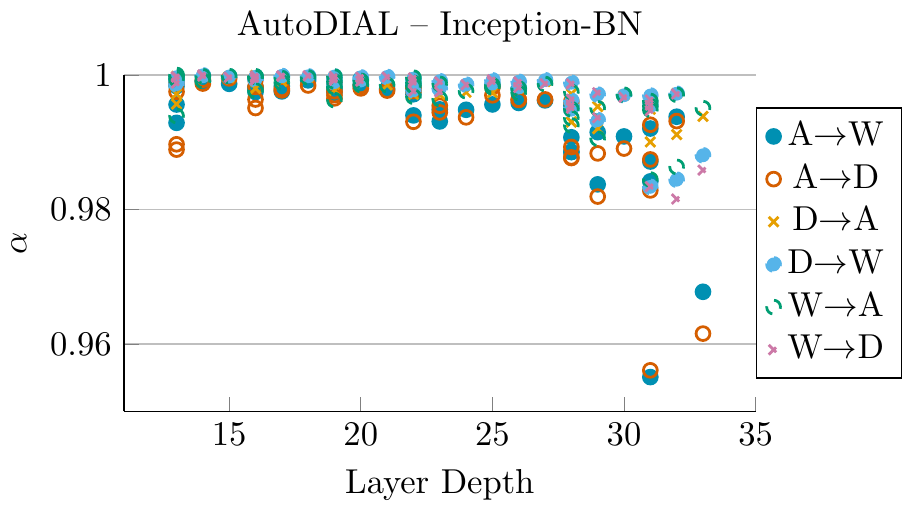}\hfill
  \includegraphics[height=3.5cm]{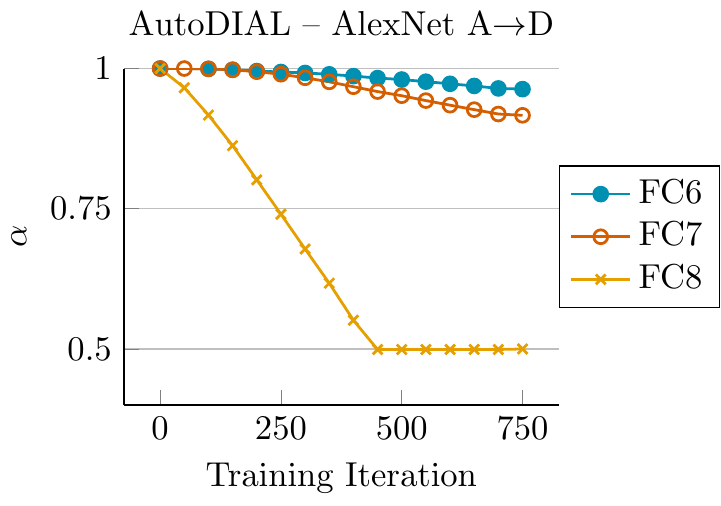}
  \caption{$\alpha$ parameters learned on the Office31 dataset, plotted as a function of layer depth (left and center) and training iteration (right).}\vspace{-10pt}
  \label{fig:alpha}
\end{figure*}

We conduct an in-depth analysis of the proposed approach, evaluating the impact of our three main contributions: i) aligning features by matching source and target distributions to a reference one; ii) learning the adaptation coefficients $\alpha$; iii) applying an entropy-based regularization term.
As a first set of experiments, we perform an ablation study on the Office31 dataset and report the results in Fig.~\ref{fig:ablation}.
Here, we compare the performance of four variations of the \Alex network: trained on source data only (Source); with the addition of the entropy loss (Entropy); with \DALs and $\alpha$ fixed to 1 (\DIAL-fixed); with \DALs and learned $\alpha$ (\DIAL).
Here the advantage of learning $\alpha$ is evident, as \DIAL outperforms \DIAL-fixed in all but one of the experimental settings.
Interestingly, the addition of the entropy term by itself seems to have mixed effects on the final accuracy: in D$\rightarrow$A and W$\rightarrow$A in particular, the performance drastically decreases in Entropy compared to Source.
This is not surprising as these two settings correspond to cases where the number of labeled source samples is very limited and the domain shift is more severe. However, using \DALs in conjunction with the entropy loss always leads to a sizable performance increase.
These results confirms the validity of our contribution: the entropy regularization term is especially beneficial when source and target data representations are aligned.

In Fig.~\ref{fig:alpha} we plot the values of $\alpha$ learned by the \DALs in \DIALAlex{} and \DIALInception{} on the Office31 dataset.
In both networks we observe that lower layers tend to learn values closer to $1$, \ie require an higher degree of adaptation compared to the layers closer to the classifier.
This behavior, however, seems to be more pronounced in \DIALAlex{} compared to \DIALInception{}.
Our results agree with recent findings in the literature \cite{aljundi2016lightweight}, according to which lower layers in a network are subject to domain shift even more than the very last layers.
During training, the $\alpha$ are able to converge to their final values in a few iterations (Fig.~\ref{fig:alpha}, right).

\subsection{Comparison with State of the Art methods}
\label{sec:results-comparison}

\begin{table*}
  \centering
  \begin{tabular}{lr*{6}{c}c}
    \toprule
    \multirow{2}{*}{Method} & \small{Source} & Amazon & Amazon & DSLR   & DSLR   & Webcam & Webcam & \multirow{2}{*}{Average}\\
                            & \small{Target} & DSLR   & Webcam & Amazon & Webcam & Amazon & DSLR   & \\
    \midrule
    \multicolumn{2}{l}{\Alex{} -- source~\cite{krizhevsky2012imagenet}}
      & $63.8$ & $61.6$ & $51.1$ & $95.4$ & $49.8$ & $99.0$ & $70.1$ \\
    \multicolumn{2}{l}{DDC~\cite{tzeng2014deep}}
      & $64.4$ & $61.8$ & $52.1$ & $95.0$ & $52.2$ & $98.5$ & $70.6$ \\
    \multicolumn{2}{l}{DAN~\cite{long2015learning}}
      & $67.0$ & $68.5$ & $54.0$ & $96.0$ & $53.1$ & $99.0$ & $72.9$ \\
    \multicolumn{2}{l}{ReverseGrad~\cite{ganin2014unsupervised}}
      & $67.1$ & $72.6$ & $54.5$ & $96.4$ & $52.7$ & $99.2$ & $72.7$ \\
                \multicolumn{2}{l}{DRCN~\cite{ghifary2016deep}}
      & $66.8$ & $68.7$ & $56.0$ & $96.4$ & $54.9$ & $99.0$ & $73.6$ \\
    \multicolumn{2}{l}{RTN~\cite{long2016unsupervised}}
      & $71.0$ & $73.3$ & $50.5$ & $\mathbf{96.8}$ & $51.0$ & $\mathbf{99.6}$ & $73.7$\\
    \multicolumn{2}{l}{JAN~\cite{long2016deep}}
      & $71.8$ & $74.9$ & $\mathbf{58.3}$ & $96.6$ & $55.0$ & ${99.5}$ & $76.0$ \\
    \midrule
    \multicolumn{2}{l}{\DIALAlex}
      & $\mathbf{73.6}$ & $\mathbf{75.5}$ & $58.1$ & $96.6$ & $\mathbf{59.4}$ & ${99.5}$ & $\mathbf{77.1}$ \\
    \bottomrule
  \end{tabular}
  \vspace{0.5em}
  \caption{AlexNet-based approaches on Office31 / full sampling protocol.}
  \label{tab:office31-alex}
\end{table*}

\begin{table*}
  \centering
  \begin{tabular}{lr*{6}{c}c}
    \toprule
    \multirow{2}{*}{Method} & \small{Source} & Amazon & Amazon & DSLR   & DSLR   & Webcam & Webcam & \multirow{2}{*}{Average}\\
                            & \small{Target} & DSLR   & Webcam & Amazon & Webcam & Amazon & DSLR   & \\
    \midrule
    \multicolumn{2}{l}{\Inception{} -- source~\cite{ioffe2015batch}}
      & $70.5$ & $70.3$ & $60.1$ & $94.3$ & $57.9$ & $\mathbf{100.0}$ & $75.5$ \\
    \multicolumn{2}{l}{AdaBN~\cite{li2016revisiting}}
      & $73.1$ & $74.2$ & $59.8$ & $95.7$ & $57.4$ & $99.8$ & $76.7$ \\
    \multicolumn{2}{l}{AdaBN + CORAL~\cite{li2016revisiting}}
      & $72.7$ & $75.4$ & $59.0$ & $96.2$ & $60.5$ & $99.6$ & $77.2$ \\
    \multicolumn{2}{l}{DDC~\cite{tzeng2014deep}}
      & $73.2$ & $72.5$ & $61.6$ & $95.5$ & $61.6$ & $98.1$ & $77.1$ \\
    \multicolumn{2}{l}{DAN~\cite{long2015learning}}
      & $74.4$ & $76.0$ & $61.5$ & $95.9$ & $60.3$ & $98.6$ & $77.8$ \\
    \multicolumn{2}{l}{JAN~\cite{long2016deep}}
      & $77.5$ & $78.1$ & $\mathbf{68.4}$ & $96.4$ & $\mathbf{65.0}$ & $99.3$ & $80.8$ \\
    \midrule
    \multicolumn{2}{l}{\DIALInception}
      & $\mathbf{82.3}$ & $\mathbf{84.2}$ & $64.6$ & $\mathbf{97.9}$ & $64.2$ & $99.9$ & $\mathbf{82.2}$ \\
    \bottomrule
  \end{tabular}
  \vspace{0.5em}
  \caption{Inception-based approaches on Office31 / full sampling protocol.}\vspace{-5pt}
  \label{tab:office31-inception}
\end{table*}

\begin{table*}[!th]
  \centering
  \begin{tabular}{lr*{6}{c}c}
  \toprule
  \multirow{2}{*}{Method} & \small{Source} & Amazon & Webcam & DSLR & Caltech & Caltech   & Caltech  & \multirow{2}{*}{Average} \\
                          & \small{Target} & Caltech & Caltech   & Caltech & Amazon & Webcam & DSLR & \\
  \midrule
  \multicolumn{2}{l}{AlexNet -- source~\cite{krizhevsky2012imagenet}} 
    & $83.8$ & $76.1$ & $80.8$ & $91.1$ & $83.1$ & $89.0$ & $84.0$\\
  \multicolumn{2}{l}{DDC~\cite{tzeng2014deep}} 
    & $85.0$ & $78.0$ & $81.1$ & $91.9$ & $85.4$ & $88.8$ & $85.0$\\
  \multicolumn{2}{l}{DAN~\cite{long2015learning}} 
    & $85.1$ & $84.3$ & $82.4$ & $92.0$ & $90.6$ & $90.5$ & $87.5$\\
  \multicolumn{2}{l}{RTN~\cite{long2016unsupervised}} 
    & $\mathbf{88.1}$ & $86.6$ & $84.6$ & $93.7$ & $\mathbf{96.9}$ & $\mathbf{94.2}$ & $\mathbf{90.6}$\\
  \midrule
  \multicolumn{2}{l}{\DIALAlex}
    & $87.4$ & $\mathbf{86.8}$ & $\mathbf{86.9}$ & $\mathbf{94.3}$ & $96.3$ & $90.1$ & $90.3$\\
  \bottomrule
  \end{tabular}
  \vspace{0.5em}
  \caption{Results on the Office-Caltech dataset using the full protocol.}\vspace{-10pt}
  \label{tab:office-caltech}
\end{table*}

\begin{table}
  \centering
  \begin{tabular}{lr*{2}{c}}
    \toprule
    \multirow{2}{*}{Method} & \small{Source} & Caltech  & Imagenet \\
                            & \small{Target} & Imagenet & Caltech  \\
    \midrule
    \multicolumn{2}{l}{SDT~\cite{tzeng2015simultaneous}}
      & -- & $73.6$ \\
    \multicolumn{2}{l}{Tommasi \etal~\cite{tommasi2016learning}}
      & -- & $75.4$ \\
    \midrule
    \multicolumn{2}{l}{\Inception{} -- source~\cite{ioffe2015batch}}
      & $82.1$ & $88.4$ \\
    \multicolumn{2}{l}{AdaBN~\cite{li2016revisiting}}
      & $82.2$  & $87.3$ \\
    \midrule
    \multicolumn{2}{l}{\DIALInception}
      & $\textbf{85.2}$ & $\textbf{90.5}$ \\
    \bottomrule
  \end{tabular}
  \vspace{1em}
  \caption{Results on the Cross-Dataset Testbed using the experimental setup in~\cite{tommasi2014testbed}.}\vspace{-18pt}
  \label{tab:ci40-default}
\end{table}

\begin{table}
  \centering
  \begin{tabular}{lr*{2}{c}}
    \toprule
    \multirow{2}{*}{Method} & \small{Source} & Caltech  & Imagenet \\
                            & \small{Target} & Imagenet & Caltech  \\
    \midrule
    \multicolumn{2}{l}{SA~\cite{fernando2013unsupervised}}
      & $43.7$ & $52.0$ \\
    \multicolumn{2}{l}{GFK~\cite{gong2012geodesic}}
      & $52.0$ & $58.5$ \\
    \multicolumn{2}{l}{TCA~\cite{pan2011domain}}
      & $48.6$ & $54.0$ \\
    \multicolumn{2}{l}{CORAL~\cite{sun2016return}}
      & $66.2$ & $74.7$ \\
    \midrule
    \multicolumn{2}{l}{\Inception{} -- source~\cite{ioffe2015batch}}
      & $82.1$ & $88.4$ \\
    \multicolumn{2}{l}{AdaBN~\cite{li2016revisiting}}
      & $81.9$  & $86.5$ \\
    \midrule
    \multicolumn{2}{l}{\DIALInception}
      & $\mathbf{84.2}$ & $\mathbf{89.8}$ \\
    \bottomrule
  \end{tabular}
  \vspace{1em}
  \caption{Results on the Cross-Dataset Testbed using the experimental setup in~\cite{sun2016return}.}\vspace{-18pt}
  \label{tab:ci40-saenko}
\end{table}

In this section we compare our approach with state-of-the art deep domain adaptation methods. We first consider the Office-31 dataset.
The results of our evaluation, obtained embedding the proposed DA-layers in the AlexNet and the Inception-BN networks as explained in Section~\ref{sec:setup}, are summarized in Tables~\ref{tab:office31-alex} and~\ref{tab:office31-inception}, respectively.
As baselines, we consider: Deep Adaptation Networks (DAN) \cite{long2015learning}, Deep Domain Confusion (DDC) \cite{tzeng2014deep}, the ReverseGrad network \cite{ganin2014unsupervised}, Residual Transfer Network (RTN) ~\cite{long2016unsupervised}, Joint Adaptation Network (JAN) \cite{long2016deep}, Deep Reconstruction-Classification Network
(DRCN) \cite{ghifary2016deep} and AdaBN~\cite{li2016revisiting} with and without CORAL feature alignment \cite{sun2016return}. The results associated to the baseline methods are derived from the original papers.
As a reference, we further report the results obtained considering standard AlexNet and Inception-BN networks trained only on source data. 

Among the deep methods based on the AlexNet architecture, \DIALAlex \ shows the best average performance, clearly demonstrating the benefit of the proposed adaptation strategy.
Similar results are found in the experiments with Inception-BN network, where our approach also outperforms all baselines.
It is interesting to compare \DIAL with the AdaBN method \cite{li2016revisiting}, as this approach also develops from a similar intuition than ours. Our results clearly demonstrate the added value of our contributions: the introduction of the alignment parameters $\alpha$, together with the adoption of the entropy regularization term, produce a significant boost in performance.

In our second set of experiments we analyze the performance of several approaches on the Office-Caltech dataset.
The results are reported in Table~\ref{tab:office-caltech}.
We restrict our attention to methods based on deep architectures and, for a fair comparison, we consider all AlexNet-based approaches.
Here we report results obtained with DDC~\cite{tzeng2014deep}, DAN~\cite{long2015learning}, and the recent Residual Transfer Network (RTN) in~\cite{long2016unsupervised}.
As it is clear from the table, our method and RTN have roughly the same performance (90.6$\%$ vs 90.4$\%$ on average), while they significantly outperform the other baselines.


Finally, we perform some experiments on the Caltech-ImageNet subset of the Cross-Dataset Testbed of \cite{tommasi2014testbed}. 
As explained above, to facilitate comparison with previous works which have also considered this dataset we perform experiments in two different settings. As baselines we consider Geodesic Flow Kernel (GFK) \cite{gong2012geodesic}, Subspace Alignment (SA) \cite{fernando2013unsupervised}), CORAL \cite{sun2016return}, Transfer Component Analysis (TCA)~\cite{pan2011domain}, Simultaneous Deep Transfer (SDT)~\cite{tzeng2015simultaneous}, and the recent method in \cite{tommasi2016learning}.
Table \ref{tab:ci40-default} and Table \ref{tab:ci40-saenko} show our results.
The proposed approach significantly outperforms previous methods and sets the new state of the art on this dataset.
The higher performance of our method is not only due to the use of Inception-BN but also due to the effectiveness of our contributions. 
Indeed, the proposed alignment strategy, combined with the adoption of the entropy regularization term, 
makes our approach more effective than previous adaptation techniques based on Inception-BN, \ie AdaBN \cite{li2016revisiting}.

\section{Conclusions}
\label{conclu}
We presented AutoDIAL, a novel framework for unsupervised, deep domain adaptation.
The core of our contribution is the introduction of novel Domain Alignment layers, which reduce the domain shift by matching source and target distributions to a reference one. Our DA-layers are endowed with a set of alignment parameters, also learned by the network, which allow the CNN not only to align the source and target feature representations but also to automatically decide at each layer the required degree of adaptation. 
Our framework exploits target data both by computing statistics in the DA-layers and by introducing an entropy loss which promotes classification models with high confidence on unlabeled samples.
The results of our experiments demonstrate that our approach outperforms state of the art domain adaptation methods.

While this paper focuses on the challenging problem of unsupervised domain-adaptation, our approach can be also exploited in a semi-supervised setting. Future works will be devoted to analyze the effectiveness of \DIAL in this scenario. Additionally, we plan to extend the proposed framework to handle multiple source domains. 

\section{Acknowledgements}
\label{sec:ack}
This work was partially founded by: project CHIST-ERA ALOOF, project ERC \#637076 RoboExNovo (F.M.C., B. C.), and project DIGIMAP, funded under grant \#860375 by the Austrian Research Promotion Agency.

\include{supp_for_main}
\FloatBarrier
{\small
\bibliographystyle{ieee}
\bibliography{egbib}
}

\end{document}

%% file: intro.tex
In spite of the progress brought by deep learning in visual recognition, the ability to generalize across different visual domains is still out of reach.
The assumption that training (source) and test (target) data are independently and identically drawn from the same distribution does not hold in many real world applications.
Indeed, it has been  shown that, even with powerful deep learning models, the domain shift problem can be alleviated but not removed \cite{donahue2014decaf}.

In the last few years the research community has devoted significant efforts in addressing domain shift.
In this context, the specific problem of unsupervised domain adaptation, \ie no labeled data are available in the target domain, deserves special attention.
In fact, in many applications annotating data is a tedious operation or may not be possible at all.
Several approaches have been proposed, both considering hand-crafted features~\cite{huang2006correcting,gong2013connecting,gong2012geodesic,long2013transfer,fernando2013unsupervised} and deep models~\cite{long2015learning,tzeng2015simultaneous,ganin2014unsupervised,long2016unsupervised,ghifary2016deep,li2016revisiting}.
In particular, recent works based on deep learning have achieved remarkable performance.
Most of these methods attempt to reduce the discrepancy among source and target distributions by learning features that are invariant to the domain shift.
Two main strategies are traditionally employed.
One is based on the minimization of Maximum Mean Discrepancy (MMD)~\cite{long2015learning,long2016unsupervised}: the distributions of the learned source and target representations are optimized to be as similar as possible by minimizing the distance between their mean embeddings.
The other strategy \cite{tzeng2015simultaneous,ganin2014unsupervised} relies on the domain-confusion loss, introduced to learn an auxiliary classifier predicting if a sample comes from the source or from the target domain.
Intuitively, by maximizing this term, \ie by imposing the auxiliary classifier to exhibit poor performance, domain-invariant features can be obtained.

\begin{figure*}[t]
  \centering
  \includegraphics[width=0.88\textwidth]{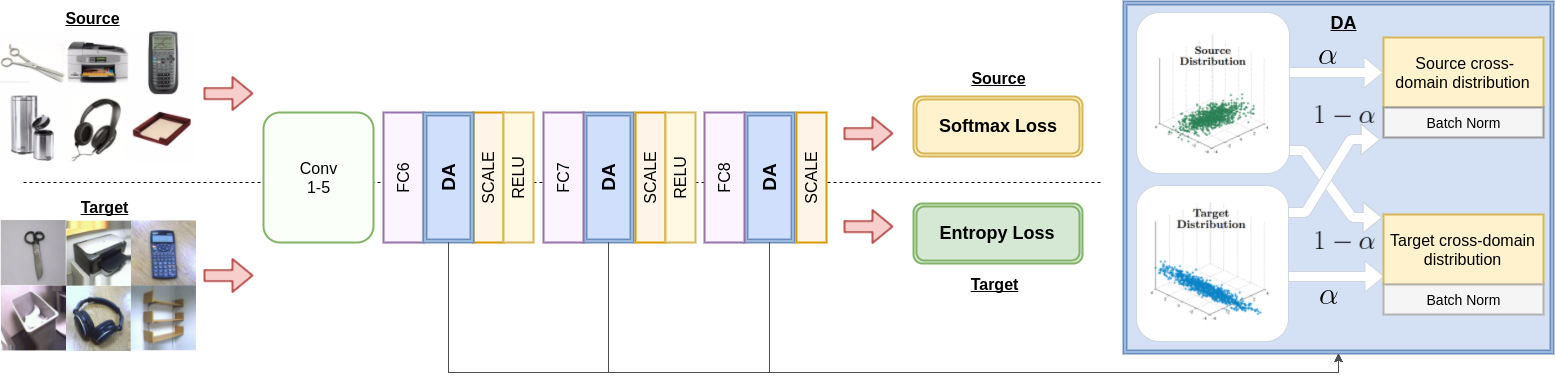}
  \caption{\DIAL as applied on \Alex~\cite{krizhevsky2012imagenet}. Source and target images are fed to the network. After passing through the first layers, they enter our \DAL where source and target distributions are aligned. The DA-layer learns the statistics of newly defined \emph{source and target cross domain distributions} and normalize the source and target mini-batches according to the computed mean and variance, different for the two domains (see Section \ref{sec:predictors}). The amount by which each distribution is influenced by the other and therefore the degree of domain alignment, depends on a parameter, $\alpha \in [0.5, 1.0]$, which is also automatically learned. After flowing through the whole network, source samples contribute to a Softmax loss, while target samples contribute to an Entropy loss, which promotes classification models which maximally separate unlabeled data. Note that we use multiple \DALs to align learned feature representations at different levels.}\vspace{-10pt}
  \label{fig:teaser}
\end{figure*}

More recently, researchers have also started to investigate alternative directions~\cite{ghifary2016deep,bousmalis2016domain,li2016revisiting,carlucci2017just}, such as the use of encoder-decoder networks to jointly learn source labels and reconstruct unsupervised target images, or the possibility of reducing the domain shift by designing specific distribution normalization layers.
In particular, the latter idea is exploited in \cite{li2016revisiting}, where a simple parameter-free approach for deep domain adaptation, called Adaptive Batch Normalization (AdaBN), is proposed.
Inspired by the popular Batch Normalization (BN) technique \cite{ioffe2015batch}, AdaBN modifies the Inception-BN network and aligns the learned source and target representations by using different mean/variance terms for the source and target domain when performing BN at test time.
This leads to learning domain-invariant features without requiring additional loss terms (\eg MMD, domain-confusion) in the optimization function and the associated extra-parameters. 

Inspired by \cite{li2016revisiting}, this paper introduces novel \textit{Domain Alignment} layers (\DALs) (Fig.\ref{fig:teaser}) which are embedded at different levels of the deep architecture to align the learned source and target feature distributions to a canonical one.
Different from \cite{li2016revisiting} and all previous deep domain adaptation methods which decide \emph{a priori} which layers should be adapted, we endow our DA-layers with the ability to \emph{automatically} learn the degree of alignment that should be pursued at different levels of the network.
This is to our knowledge the first work that tries to pursue this objective.
Furthermore, we argue that in \cite{li2016revisiting} unlabeled target data are not fully exploited (see Sec.~\ref{sec:predictors}).
Instead, we leverage information from the target domain to construct a prior distribution on the network parameters, biasing the learned solution towards models that are able to separate well the classes in the target domain (see Sec.~\ref{ss:training} and~\cite{grandvalet2004semisupervised}).
Our DA-layers and the considered prior distribution work in synergy during learning: the first aligning the source and target feature distributions, the second encouraging the network to learn features that lead to maximally separated target classes.
We call our algorithm AutoDIAL -- DomaIn Alignment Layers with Automatic alignment parameters.
An extensive experimental evaluation demonstrates that AutoDIAL greatly alleviates the domain discrepancy and outperforms state of the art techniques on three popular benchmarks: Office-31~\cite{saenko2010adapting}, Office-Caltech~\cite{gong2012geodesic} and the Caltech-ImageNet setting of the Cross-Dataset Testbed\cite{tommasi2014testbed}.

\textbf{Contributions}.
The contribution of this work is threefold.
First, we present an approach for unsupervised domain adaptation, based on the introduction of DA-layers to explicitly address the domain shift problem, which act in synergy with an entropy loss which exploits unsupervised target data during learning.
Our solution simultaneously aligns feature representations and learns where and to which extent adaptation should take place.
Second, in contrast to previous works optimizing domain discrepancy regularization terms~\cite{long2016unsupervised,tzeng2015simultaneous,ganin2014unsupervised,long2015learning}, our DA-layers do not require any additional meta-parameters.
Third, we perform an extensive experimental analysis on three different benchmarks.
We find that our unsupervised domain adaptation approach outperforms state-of-the-art methods and can be applied to different CNN architectures, consistently improving their performance in domain adaptation problems.

%% file: related.tex
Unsupervised domain adaptation focuses on the 
scenario where
labeled data are only available in the source domain. Traditional methods 
addressed the problem of reducing the discrepancy between the source
and the target distributions by considering two main strategies. 
The first
is based on instance re-weighting \cite{huang2006correcting,chu2013selective,yamada2012no,gong2013connecting,zeng2014deep}. Initially, 
source samples are assigned different importance according to their similarity with the target data. Then, the re-weighted instances are used
to learn a classification/regression model for the target domain. Following this scheme, Huang \etal\cite{huang2006correcting}
introduced Kernel Mean Matching, a nonparametric
method to set source sample weights without explicitly estimating the data distributions.
Gong \etal\cite{gong2013connecting} proposed 
to automatically discover landmark datapoints, \ie
the subset of source instances being more similar to target data, and used them to create
domain-invariant features. Chu \etal \cite{chu2013selective} formalized the two tasks of sample selection
and classifier learning within a single optimization problem. 
While these works considered hand-crafted features, recently similar ideas have been
applied to deep models. For instance, Zeng \etal \cite{zeng2014deep} described 
an unsupervised domain adaptation approach for pedestrian detection 
using deep autoencoders to weight the importance of source training samples.

A second strategy for unsupervised domain adaptation
is based on feature alignment, \ie source and target data are projected
in a common subspace as to reduce the distance among the
associated distributions. This approach 
attracted 
considerable interest in the past years and several different methods have been 
proposed, both considering shallow models \cite{gong2012geodesic,long2013transfer,fernando2013unsupervised} and deep architectures 
\cite{long2015learning,tzeng2015simultaneous,ganin2014unsupervised,ghifary2016deep,bousmalis2016domain}.
Focusing on recent deep domain adaptation methods,
two different schemes are typically considered for aligning feature representations: (i)
multiple adaptation schemes are introduced in order to reduce Maximum Mean Discrepancy \cite{long2015learning,long2016unsupervised,sun2016deep} or (ii) deep features are learned
in a domain-adversarial setting, \ie maximizing a domain confusion loss \cite{tzeng2015simultaneous,ganin2014unsupervised}.
Our approach belongs to the category of methods employing deep learning for domain adaptation. However,
we significantly depart from previous works, reducing the discrepancy between 
source and target distributions by introducing a domain alignment approach based on DA-layers. The closest work to ours is
\cite{li2016revisiting}, where Li \etal propose to use BN in the context of domain adaptation. Our approach 
can be seen as a generalization of \cite{li2016revisiting}, as our DA layers allows to automatically tune the required degree of adaptation at each level of the deep network. Furthermore, we also introduce a prior over the network parameters in order to fully benefit from the target samples during training.
Experiments presented in Section~\ref{experiments} show the significant added value of our idea.

%% file: method.tex
\label{method}
Let $\set X$ be the input space (\eg images) and $\set Y$ the output space (\eg image categories) of our learning task.
In unsupervised domain adaptation, we have a \emph{source} domain and a \emph{target} domain that
are identified via probability distributions $p^s_{\mathtt{xy}}$ and $p^t_\mathtt{xy}$, respectively, defined over $\set X\times\set Y$.
The two distributions are in general different and unknown, but we are provided with a source dataset $\set S=\{(x^s_1,y^s_1),\dots,(x^s_n,y^s_n)\}$ of \emph{i.i.d.} observations from $p^s_{\mathtt{xy}}$ and an unlabeled target dataset $\set T=\{x^t_1,\dots,x^t_m\}$ of \emph{i.i.d.} observations from the marginal $p^t_\mathtt x$.
The goal is to estimate a predictor from $\set S$ and $\set T$ that can be used to classify sample points from the target domain.
This task is particularly challenging because on one hand we lack direct observations of labels from the target domain and on the other hand the discrepancy between the source and target domain distributions prevents a predictor trained on $\set S$ to be readily applied to the target domain. 

A number of state of the art methods try to reduce the domain discrepancy by performing some form of alignment at the feature or classifier level.
In particular, the recent, most successful methods try to \emph{couple} the training process and the domain adaptation step within \emph{deep} neural architectures~\cite{ganin2014unsupervised,long2016unsupervised,long2015learning}, as this solution enables alignments at different levels of abstraction.
The approach we propose embraces the same philosophy, while departing from the assumption that domain alignment can by pursued by applying the \emph{same} predictor to the source and target domains.
This is motivated by an impossibility theorem~\cite{ben2010impossibility}, which intuitively states that no learner relying on the \emph{covariate shift} hypothesis, \ie $p^s_{\mathtt y|\mathtt x}=p^t_{\mathtt y|\mathtt x}$, and achieving a low discrepancy between the source and target unlabeled distributions $p^s_\mathtt x$ and $p^t_\mathtt x$, is guaranteed to succeed in domain adaptation without further relatedness assumptions between training and target distributions.
For this reason, we assume that the source and target predictors are in general \emph{different} functions.
Nonetheless, both predictors depend on a common parameter $\theta$ belonging to a set $\Theta$, which couples explicitly the two predictors, while not being directly involved in the alignment of the source and target domains.
This contrasts with the majority of state of the art methods that augment the loss function used to train their predictors with a regularization term penalizing discrepancies between source and target representations (see, \eg~\cite{ganin2014unsupervised,long2016unsupervised,long2015learning}). 
The perspective we take is different and is close in spirit to AdaBN~\cite{li2016revisiting}.
It consists in hard-coding the desired domain-invariance properties into the source and target predictors through the introduction of so-called \emph{Domain-Alignment layers} (\DALs).
Moreover, we sidestep the problem of deciding which layers should be aligned, and to what extent, by endowing the architecture with the ability to \emph{automatically} tune the degree of alignment that should be considered in each domain-alignment layer.
The rest of this section is devoted to providing the details of our method.

\subsection{Source and Target Predictors}
\label{sec:predictors}
The source and target predictors are implemented as two deep neural networks being almost identical, as they share the same structure and the same weights (given by the parameter $\theta$).
However, the two networks contain also a number of special layers, the \DALs, which implement a domain-specific operation.
Indeed, the role of such layers is to apply a data transformation that aligns the observed input distribution with a reference distribution.
Since in general the input distributions of the source and target predictors differ, while the reference distribution stays the same, we have that the two predictors undergo different transformations in the corresponding \DALs.
Consequently, the source and target predictors de facto implement different functions, which is important for the reasons given in Sec.~\ref{method}.

The actual implementation of our \DALs is inspired by AdaBN~\cite{li2016revisiting}, where Batch Normalization layers are used to independently align source and target distributions to a standard normal distribution, by matching the first- and second-order moments.
The approach they propose consists in training on the source a network having BN-layers, thus obtaining the source predictor, and deriving the target predictor as a post-processing step, which re-estimates the BN statistics using target samples only.
Accordingly, the source and target predictors share the same network parameters but have different BN statistics, thus rendering the two predictors different functions.

The approach we propose sticks to the same idea of using BN-layers to align domains, but we introduce fundamental changes.
One limitation of AdaBN is that the target samples have no influence on the network parameters, as they are not observed during training.
Our approach overcomes this limitation by coupling the network parameters to both target and source samples at training time.
This is achieved in two ways: first we introduce a prior distribution for the network parameters based on the target samples; second, we endow the architecture with the ability of learning the degree of adaptation by introducing a parametrized, cross-domain bias to the input distribution of each domain-specific \DAL.
The rest of this subsection is devoted to describe the new layer, while we defer to the next subsection the description of the prior distribution.

\paragraph{\DAL.}
As mentioned before, our \DAL is derived from Batch Normalization, but as opposed to BN, which computes first and second-order moments from the input distribution derived from the mini-batch, we let the latter statistics to be contaminated by samples from the other domain, thus introducing a cross-domain bias. 
Since the source and target predictors share the same network topology, each \DAL in one predictor has a matching \DAL in the other predictor.
Let $x_s$ and $x_t$ denote inputs to matching \DALs in the source and target predictor, respectively, for a given feature channel and spatial location.
Assume $q^s$ and $q^t$ to be the distribution of $x_s$ and $x_t$, respectively, and let $q_\alpha^{st}=\alpha q^s + (1-\alpha) q^t$ and, symmetrically, $q_\alpha^{ts}=\alpha q^t + (1-\alpha) q^s$ be cross-domain distributions mixed by a factor $\alpha\in[0.5,1]$.
Then, the output of the \DALs in the source and target networks are given respectively by
\begin{equation}
\label{eqn:dal}
\mathtt {DA}(x_s;\alpha) = \frac{x_s-\mu_{st,\alpha}}{\sqrt{\epsilon+\sigma_{st,\alpha}^2}},\,\,\,
\mathtt {DA}(x_t;\alpha) = \frac{x_t-\mu_{ts,\alpha}}{\sqrt{\epsilon+\sigma_{ts,\alpha}^2}},
\end{equation}
where $\epsilon>0$ is a small number to avoid numerical issues in case of zero variance, $\mu_{st,\alpha}=\mathtt {E}_{x\sim q_\alpha^{st}}[x]$, $\sigma^2_{st,\alpha}=\mathtt {Var}_{x\sim q_\alpha^{st}}[x]$, and similarly $\mu_{ts,\alpha}$ and $\sigma^2_{st,\alpha}$ are mean and variance of $x\sim q^{ts}_\alpha$.
Akin to BN, we estimate the statistics based on the mini-batch and derive similarly the gradients through the statistics (see Supplementary Material).

The rationale behind the introduction of the mixing factor $\alpha$ is that we can move from having an independent alignment of the two domains akin to AdaBN, when $\alpha=1$, to having a coupled normalization when $\alpha=0.5$.
In the former case the \DAL computes two different functions in the source and target predictors and is equivalent to considering a full degree of domain alignment.
The latter case, instead, yields the same function since $q_{0.5}^{st}=q_{0.5}^{ts}$ thus transforming the two domains equally, which yields no domain alignment.
Since the mixing parameter $\alpha$ is not fixed a priori but learned during the training phase, we obtain as a result that the network can decide how strong the domain alignment should be at each level of the architecture where \DAL is applied.
More details about the actual CNN architectures used to implement the two domain predictors are given in Section~\ref{sec:setup}.

\subsection{Training}\label{ss:training}
During the training phase we estimate the parameter $\theta$, which holds the neural network weights shared by the source and target predictors including the mixing factors pertaining to the \DALs, using the observations provided by the source dataset $\set S$ and the target dataset $\set T$. 
As we stick to a discriminative model, the unlabeled target dataset cannot be employed to express the data likelihood. However, we can exploit $\set T$ to construct a prior
distribution of the parameter $\theta$. Accordingly, we shape a posterior distribution of $\theta$ given the observations $\set S$ and $\set T$ as
\begin{equation}
    \pi(\theta|\set S,\set T)\propto\pi(y_{\set S}|x_{\set S},\set T,\theta)\pi(\theta|\set T, x_\set S)\,,
    \label{eq:posterior}
\end{equation}
where $y_{\set S}=\{y^s_1,\dots,y^s_n\}$ and $x_{\set S}=\{x^s_1,\dots,x^s_n\}$ collect the labels and data points of the observations in $\set S$, respectively.
The posterior distribution is maximized over $\Theta$ to obtain a maximum a posteriori estimate $\hat\theta$ of the parameter used in the source and target predictors:
\begin{equation}
    \hat\theta \in\argmax_{\theta\in\Theta} \pi(\theta|\set S,\set T)\,.
    \label{eq:MAP}
\end{equation}
The term $\pi(y_{\set S}|x_{\set S},\set T,\theta)$ in~\eqref{eq:posterior} represents the likelihood of $\theta$ with respect to the source dataset,
while
$\pi(\theta|\set T, x_\set S)$ is the prior term depending on the target dataset, which acts as a regularizer in the classical learning theory sense. Both terms actually, depend on both domains due to the cross-domain statistics that we have in our \DALs for $\frac{1}{2}\leq\alpha<1$ and are estimated from samples from the source \emph{and} target domains.

The likelihood decomposes into the following product over sample points, due to the data sample \emph{i.i.d.} assumption:
\begin{equation}
    \pi(y_{\set S}|x_{\set S},\set T,\theta)=\prod_{i=1}^nf_s^\theta(y^s_i;x^s_i)\,,
    \label{eq:likelihood}
\end{equation}
where $f^\theta_s(y^s_i;x^s_i)$ is the probability that sample point $x^s_i$ takes label $y^s_i$ according to the source predictor (we omitted the dependence on $\set T$ and $x_\set S$ for notational convenience).

Before delving into the details of the prior term, we would like to remark on the absence of an explicit component in the probabilistic model that tries to align the source and target distributions.
This is because in our model the domain-alignment step is taken over by each predictor, independently, via the domain-alignment layers as shown in the previous subsection.

\paragraph{Prior distribution.}
The prior distribution of the parameter $\theta$ shared by the source and target predictors is constructed from the observed, target data distribution. This choice is motivated by the theoretical possibility of squeezing more bits of information from unlabeled data points insofar as they exhibit low levels of class overlap~\cite{oneill1978normal}. Accordingly, it is reasonable to bias a priori a predictor based on the degree of label uncertainty that is observed when the same predictor is applied to the target samples. Uncertainty in this sense can be measured for an hypothesis $\theta$ in terms of the empirical entropy of $\mathtt y|\theta$ conditioned on $\mathtt x$ as follows
\begin{equation}
    h(\theta|\set T,x_\set S)=-\frac{1}{m}\sum_{i=1}^m\sum_{y\in\set Y}f^\theta_t(y;x^t_i)\log f^\theta_t(y;x^t_i)\,,
    \label{eq:h}
\end{equation}
where $f_t(y;x^t_i)$ represents the probability that sample point $x_i^t$ takes label $y$ according to the target predictor (again we omitted the dependence on $\set T$ and $x_\set S$).

It is now possible to derive a prior distribution $\pi(\theta|\set T,x_\set S)$ in terms of the label uncertainty measure $h(\theta|\set T,x_\set S)$
by requiring the prior distribution to maximize the entropy under the constraint $\int h(\theta|\set T,x_\set S)\pi(\theta|\set T,x_\set S)d\theta=\varepsilon$, 
where the constant $\varepsilon>0$ specifies how small the label uncertainty should be on average. This yields a concave, variational optimization problem with solution:
\begin{equation}
    \pi(\theta|\set T,x_\set S)\propto\exp\left( -\lambda\, h(\theta|\set T,x_\set S) \right)\,,
    \label{eq:prior}
\end{equation}
where $\lambda$ is the Lagrange multiplier corresponding to $\varepsilon$. 
The resulting prior distribution  
satisfies the desired property of preferring models that exhibit well separated classes (\ie having lower values of $h(\theta|\set T,x_\set S)$), thus enabling 
the exploitation of the information content of unlabeled target observations within a discriminative setting~\cite{grandvalet2004semisupervised}.

Prior distributions of this kind have been adopted also in other works~\cite{long2016unsupervised} in order to exploit more information from the target distribution, but has never been used before in conjunction to explicit domain alignment methods (\ie not based on additional regularization terms such as MMD and domain-confusion) like the one we are proposing.

\paragraph{Inference.}
Once we have estimated the optimal network parameters $\hat\theta$ by solving~\eqref{eq:MAP}, we can remove the dependence of the target predictor on $\set T$ and $x_\set S$.
In fact, after fixing $\hat\theta$, the input distribution to each \DAL also becomes fixed, and we can thus compute and store the required statistics once at all, akin to standard BN. 

\subsection{Implementation Notes}
\DAL can be implemented as a mostly straightforward modification of standard Batch Normalization.
We refer the reader to the supplementary material for a complete derivation.
In our implementation in particular, we treat each pair of \DALs as a single network layer which simultaneously computes the two normalization functions in Equation~\eqref{eqn:dal} and learns the $\alpha$ parameter.
During training each batch contains a fixed number of source samples, followed by a fixed number of target samples, allowing our \DALs to simply differentiate between the two.
Similarly to standard BN, we keep separate moving averages of the source and target statistics.
Note that, as mentioned before, $\alpha \in [0.5,1]$.
We enforce this by clipping its value in the allowed range in each forward pass of the network.

By replacing the optimization problem in~\eqref{eq:MAP} with the equivalent minimization of the negative logarithm of $\pi(\theta|\set S, \set T)$ and combining \eqref{eq:posterior}, \eqref{eq:likelihood}, \eqref{eq:h} and \eqref{eq:prior} we obtain a loss function $L(\theta)=L^s(\theta) + \lambda L^t(\theta)$, where:
\begin{align*}
\label{eq:loss}
  L^s(\theta) &= - \frac{1}{n}\sum_{i=1}^n \log f_s^\theta(y_i^s;x_i^s)\,,\\
  L^t(\theta) &= - \frac{1}{m}\sum_{i=1}^m\sum_{y\in\set Y}f_t^\theta(y;x^t_i)\log f_t^\theta(y;x^t_i)\,.
\end{align*}
The term $L^s(\theta)$ is the standard log-loss applied to the source samples, while $L^t(\theta)$ is an entropy loss applied to the target samples.
The second term can be implemented by feeding $f_t^\theta(y;x^t_i)$ to both inputs of a cross-entropy loss layer, where supported by the deep learning toolkit of choice.
In our implementation, based on Caffe~\cite{jia2014caffe}, we obtain it by slightly modifying the existing \texttt{SoftmaxLoss} layer.\footnote{The source code is available at \url{https://github.com/ducksoup/autodial}}

%% file: supp_for_main.tex
\begin{appendices}
\section*{Appendix}

This document provides the following additional contributions to our ICCV 2017 submission:
\begin{itemize}
\item in Section~\ref{sec:grad}, we provide explicit formulas for the batch statistics in our \DALs as well as the layer's back-propagation equations;
\item in Section~\ref{sec:svhn}, we provide results on the SVHN -- MNIST benchmark.
\item in Section~\ref{sec:distr}, we provide some examples of feature distributions learned by \DIALInception{} on the Office 31 dataset.
\end{itemize}

\section{\DALs formulas}
\label{sec:grad}

We rewrite Eq.~(1) of the main paper to make sample indexes explicit:
\begin{equation}
\begin{aligned}
  y^s_i &= \texttt{DA}(x^s_i;\alpha) =
    \frac{x^s_i - \mu_{st,\alpha}}{\sqrt{\epsilon + \sigma_{st,\alpha}^2}}\,,\\
  y^t_i &= \texttt{DA}(x^t_i;\alpha) =
    \frac{x^t_i - \mu_{ts,\alpha}}{\sqrt{\epsilon + \sigma_{ts,\alpha}^2}}\,.
\end{aligned}
\end{equation}
Using this notation, the batch statistics are
\begin{equation}
\begin{aligned}
  \mu_{st,\alpha} &= \frac{\alpha}{\con{n}_s} \sum_{i=1}^{\con{n}_s} x^s_i +
    \frac{1-\alpha}{\con{n}_t} \sum_{i=1}^{\con{n}_t} x^t_i\,,\\
  \mu_{ts,\alpha} &= \frac{1-\alpha}{\con{n}_s} \sum_{i=1}^{\con{n}_s} x^s_i +
    \frac{\alpha}{\con{n}_t} \sum_{i=1}^{\con{n}_t} x^t_i\,,\\
  \sigma_{st,\alpha}^2 &= \frac{\alpha}{\con{n}_s} \sum_{i=1}^{\con{n}_s} (x^s_i - \mu_{st,\alpha})^2 +
    \frac{1-\alpha}{\con{n}_t} \sum_{i=1}^{\con{n}_t} (x^t_i - \mu_{st,\alpha})^2\,,\\
  \sigma_{ts,\alpha}^2 &= \frac{1-\alpha}{\con{n}_s} \sum_{i=1}^{\con{n}_s} (x^s_i - \mu_{ts,\alpha})^2 +
    \frac{\alpha}{\con{n}_t} \sum_{i=1}^{\con{n}_t} (x^t_i - \mu_{ts,\alpha})^2\,,\\
\end{aligned}
\end{equation}
where $\con{n}_s$ and $\con{n}_t$ are, respectively, the number of source and target samples in a batch.
The partial derivatives of the statistics w.r.t. the inputs are
\begin{equation}
\begin{aligned}
  \frac{\partial \mu_{st,\alpha}}{\partial x^s_j} &= \frac{\alpha}{\con{n}_s}\,, &
  \frac{\partial \mu_{st,\alpha}}{\partial x^t_j} &= \frac{1-\alpha}{\con{n}_t}\,, \\
  \frac{\partial \mu_{ts,\alpha}}{\partial x^s_j} &= \frac{1-\alpha}{\con{n}_s}\,, &
  \frac{\partial \mu_{ts,\alpha}}{\partial x^t_j} &= \frac{\alpha}{\con{n}_t}\,,
\end{aligned}
\end{equation}
\begin{equation}
\begin{aligned}
  \frac{\partial \sigma_{st,\alpha}}{\partial x^s_j} &= 2 \frac{\alpha}{\con{n}_s} (x^s_j - \mu_{st,\alpha})\,, \\
  \frac{\partial \sigma_{st,\alpha}}{\partial x^t_j} &= 2 \frac{1-\alpha}{\con{n}_t} (x^t_j - \mu_{st,\alpha})\,, \\
  \frac{\partial \sigma_{ts,\alpha}}{\partial x^s_j} &= 2 \frac{1-\alpha}{\con{n}_s} (x^s_j - \mu_{ts,\alpha})\,, \\
  \frac{\partial \sigma_{ts,\alpha}}{\partial x^t_j} &= 2 \frac{\alpha}{\con{n}_t} (x^t_j - \mu_{ts,\alpha})\,. \\
\end{aligned}
\end{equation}
The partial derivatives of the loss $L$ w.r.t. the inputs are
\begin{equation}
\label{eqn:grad}
\begin{aligned}
  \frac{\partial L}{\partial x^s_j}
  &=\frac{1}{\sqrt{\epsilon + \sigma_{st,\alpha}^2}} \left[
    \frac{\partial L}{\partial y^s_j} - \frac{\alpha}{\con{n}_s} \left(
      \sum_{i=1}^{\con{n}_s} \frac{\partial L}{\partial y^s_i} +
      y^s_j \sum_{i=1}^{\con{n}_s} y^s_i \frac{\partial L}{\partial y^s_i} \right) \right] \\
  &-\frac{1}{\sqrt{\epsilon + \sigma_{ts,\alpha}^2}} \frac{1-\alpha}{\con{n}_s} \left(
      \sum_{i=1}^{\con{n}_t} \frac{\partial L}{\partial y^t_i} +
      y^{st}_j \sum_{i=1}^{\con{n}_t} y^t_i \frac{\partial L}{\partial y^t_i} \right)\,, \\
  \frac{\partial L}{\partial x^t_j}
  &=\frac{1}{\sqrt{\epsilon + \sigma_{ts,\alpha}^2}} \left[
    \frac{\partial L}{\partial y^t_j} - \frac{\alpha}{\con{n}_t} \left(
      \sum_{i=1}^{\con{n}_t} \frac{\partial L}{\partial y^t_i} +
      y^t_j \sum_{i=1}^{\con{n}_t} y^t_i \frac{\partial L}{\partial y^t_i} \right) \right] \\
  &-\frac{1}{\sqrt{\epsilon + \sigma_{ts,\alpha}^2}} \frac{1-\alpha}{\con{n}_t} \left(
      \sum_{i=1}^{\con{n}_s} \frac{\partial L}{\partial y^s_i} +
      y^{ts}_j \sum_{i=1}^{\con{n}_s} y^s_i \frac{\partial L}{\partial y^s_i} \right)\,, \\
\end{aligned}
\end{equation}
where $y^{st}_i$ and $y^{ts}_i$ are ``cross-normalized'' outputs
\begin{equation}
\begin{aligned}
  y^{st}_i &=
    \frac{x^s_i - \mu_{ts,\alpha}}{\sqrt{\epsilon + \sigma_{ts,\alpha}^2}}\,, &
  y^{ts}_i &=
    \frac{x^t_i - \mu_{st,\alpha}}{\sqrt{\epsilon + \sigma_{st,\alpha}^2}}\,.
\end{aligned}
\end{equation}
Using these definitions, one can also compute the partial derivative of $L$ w.r.t. the domain mixing parameter $\alpha$ as
\begin{equation}
\begin{aligned}
  \frac{\partial L}{\partial\alpha}
  &=\left(
      \frac{1}{\con{n}_t} \sum_{i=1}^{\con{n}_t} y_i^{ts} -
      \frac{1}{\con{n}_s} \sum_{i=1}^{\con{n}_s} y_i^s \right)
    \sum_{i=1}^{\con{n}_s} \frac{\partial L}{\partial y_i^s}\\
  &+\left(
      \frac{1}{\con{n}_t} \sum_{i=1}^{\con{n}_t} (y_i^{ts})^2 -
      \frac{1}{\con{n}_s} \sum_{i=1}^{\con{n}_s} (y_i^s)^2 \right)
    \frac{1}{2} \sum_{i=1}^{\con{n}_s} y_i^s \frac{\partial L}{\partial y_i^s} \\
  &+\left(
      \frac{1}{\con{n}_s} \sum_{i=1}^{\con{n}_s} y_i^{st} -
      \frac{1}{\con{n}_t} \sum_{i=1}^{\con{n}_t} y_i^t \right)
    \sum_{i=1}^{\con{n}_t} \frac{\partial L}{\partial y_i^t} \\
  &+\left(
      \frac{1}{\con{n}_s} \sum_{i=1}^{\con{n}_s} (y_i^{st})^2 -
      \frac{1}{\con{n}_t} \sum_{i=1}^{\con{n}_t} (y_i^t)^2 \right)
    \frac{1}{2} \sum_{i=1}^{\con{n}_t} y_i^t \frac{\partial L}{\partial y_i^t}\,.
\end{aligned}
\end{equation}

Note that, as for standard Batch Normalization, the gradients do not depend on the layer's inputs, allowing for in-place computation where permitted by the deep learning framework of choice.

\section{Results on the SVHN -- MNIST benchmark}
\label{sec:svhn}

In this section we report the results we obtain in the SVHN~\cite{netzer2011reading} to MNIST~\cite{lecun1998gradient} transfer benchmark.
We follow the experimental protocol in~\cite{ganin2016domain}, using all SVHN images as the source domain and all MNIST images as the target domain, and compare with the following baselines:  CORAL \cite{sun2016return}; the Deep Adaptation Networks (DAN) \cite{long2015learning}; the Domain-Adversarial Neural Network (DANN) in~\cite{ganin2016domain}; the Deep Reconstruction Classification Network (DRCN) in~\cite{ghifary2016deep}; the Domain Separation Networks (DSN) in~\cite{bousmalis2016domain}; the Asymmetric Tri-training Network (ATN) in~\cite{saito2017asymmetric}.

As in all baselines, we adopt the network architecture in~\cite{ganin2014unsupervised}, adding \DALs after each layer with parameters.
Training is performed from scratch, using the same meta-parameters as for \Alex (see Main Paper), with the following exceptions: initial learning rate $l_0=0.01$; 25 epochs; learning rate schedule defined by $l_p = l_0 / (1 + \gamma p)^\beta$, where $\gamma=10$, $\beta=0.75$ and $p$ is the learning progress linearly increasing from $0$ to $1$.

\begin{figure}
  \centering
  \includegraphics[width=0.8\columnwidth]{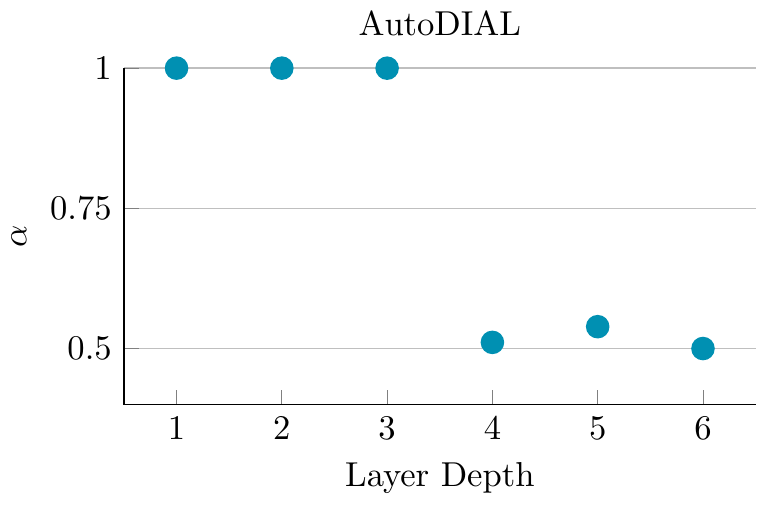}
  \caption{$\alpha$ parameters learned on the SVHN -- MNIST dataset, plotted as a function of layer depth.}
  \label{fig:alpha-svhn}
\end{figure}

As shown in Table~\ref{tab:svhn}, we set the new state of the art on this benchmark.
It is worth of note that \DIAL also outperforms the methods, such as ATN and DSN, which expand the capacity of the original network by adding numerous learnable parameters, while only employing a single extra learnable parameter in each \DAL.
The $\alpha$ parameters learned by \DIAL on this dataset are plotted in Fig.~\ref{fig:alpha-svhn}.
Similarly to the case of \Alex and \Inception on the Office-31 dataset, the network learns higher values of $\alpha$ in the bottom of the network and lower values of $\alpha$ in the top.
In this case, however, we observe a steeper transition from $1$ to $0.5$, which interestingly corresponds with the transition from convolutional to fully-connected layers in the network.

\begin{table}
  \centering
  \begin{tabular}{lc}
    \toprule
    Method & Accuracy \\
    \midrule
        CORAL~\cite{li2016revisiting} & $63.1$ \\
        DAN~\cite{long2015learning} & $71.1$ \\
    DANN~\cite{ganin2016domain} & $73.9$ \\
    DRCN~\cite{ghifary2016deep} & $82.0$ \\
    DSN~\cite{bousmalis2016domain} & $82.7$ \\
    ATN~\cite{saito2017asymmetric} & $86.2$ \\
    \midrule
    \DIAL & $\mathbf{90.3}$ \\
    \bottomrule
  \end{tabular}
  \vspace{0.5em}
  \caption{Results on the SVHN to MNIST benchmark.}
  \label{tab:svhn}
\end{table}

\section{Feature distributions}
\label{sec:distr}

\begin{figure*}[ht!]
  \includegraphics[width=0.24\textwidth]{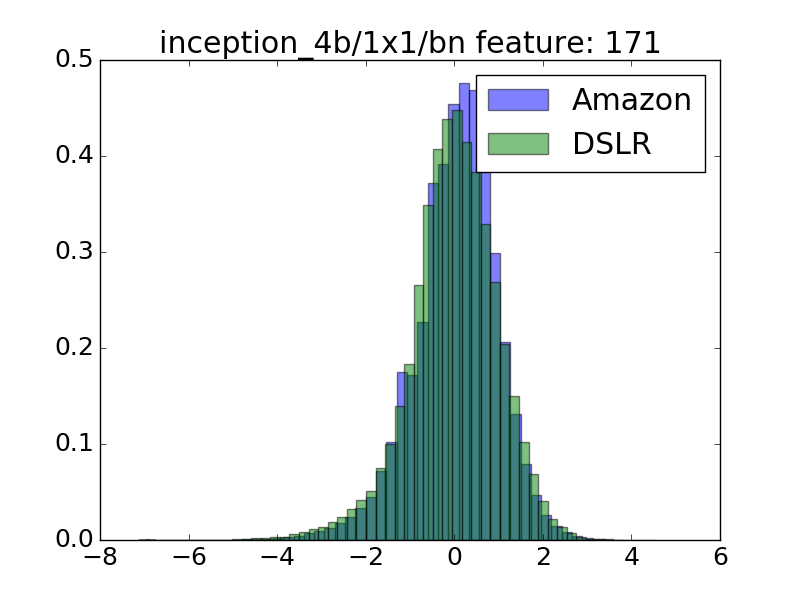}
  \includegraphics[width=0.24\textwidth]{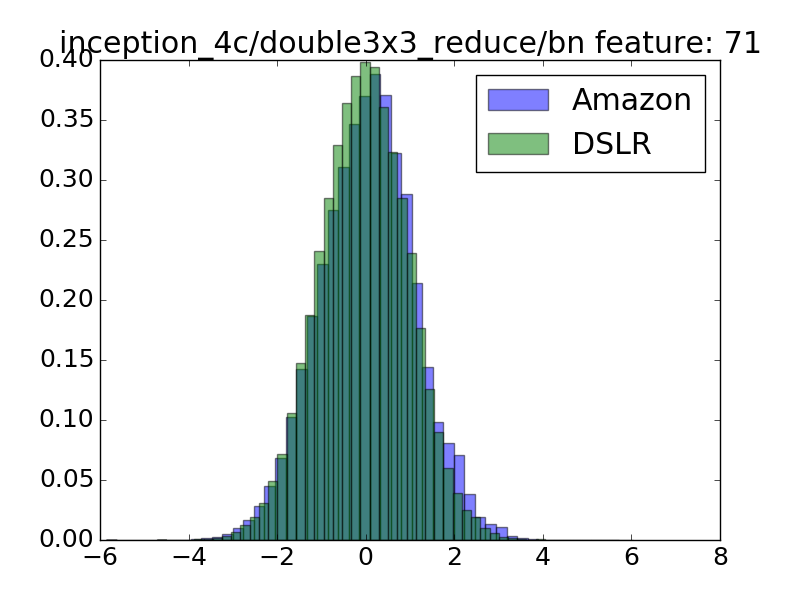}
  \includegraphics[width=0.24\textwidth]{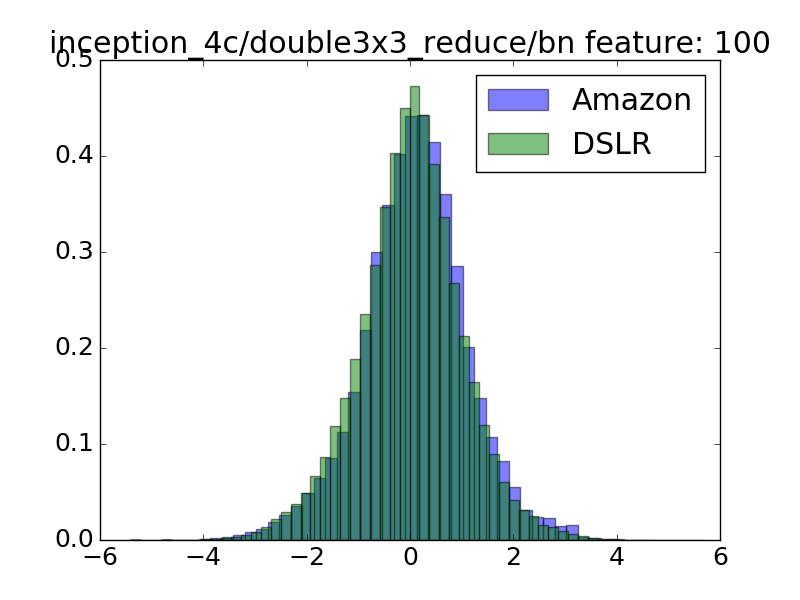}
  \includegraphics[width=0.24\textwidth]{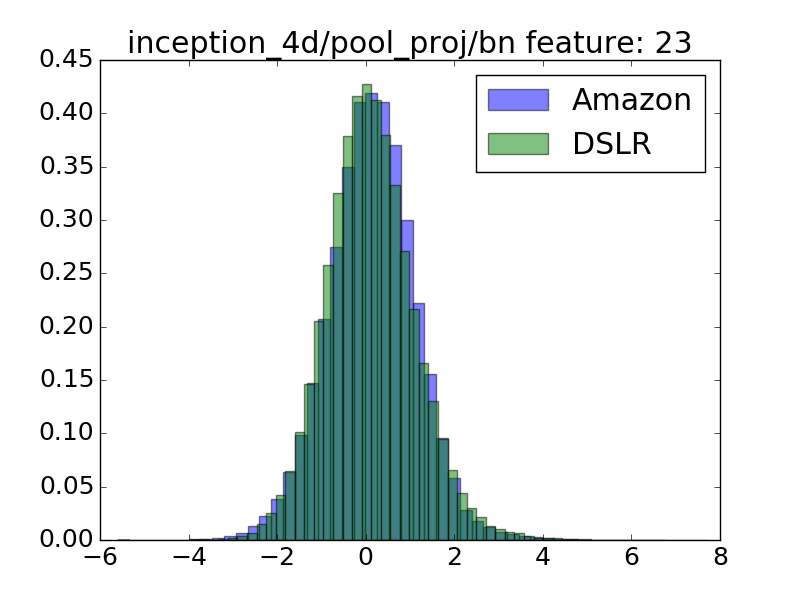} \\
  \includegraphics[width=0.24\textwidth]{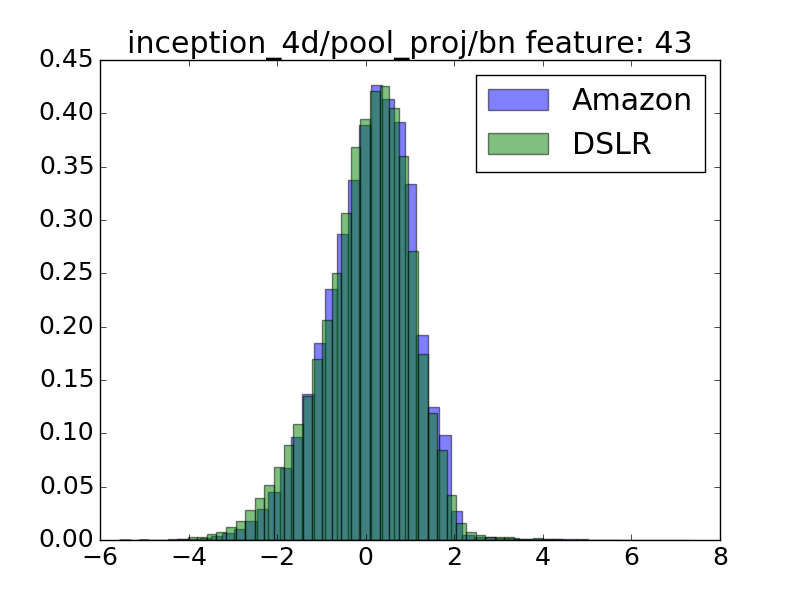}
  \includegraphics[width=0.24\textwidth]{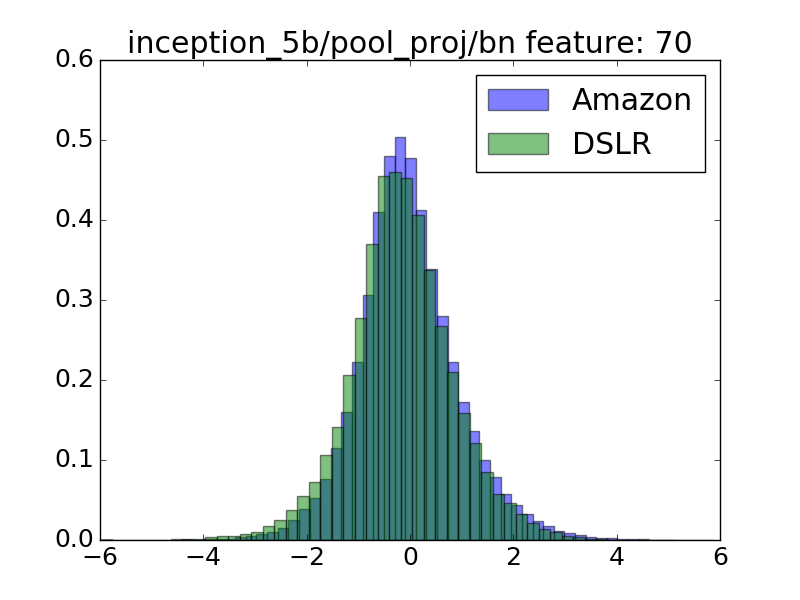}
  \includegraphics[width=0.24\textwidth]{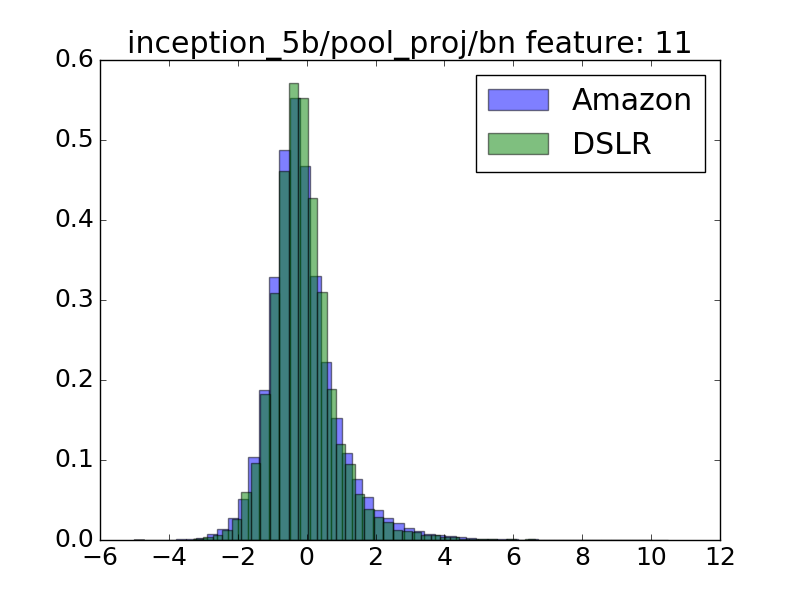}
  \includegraphics[width=0.24\textwidth]{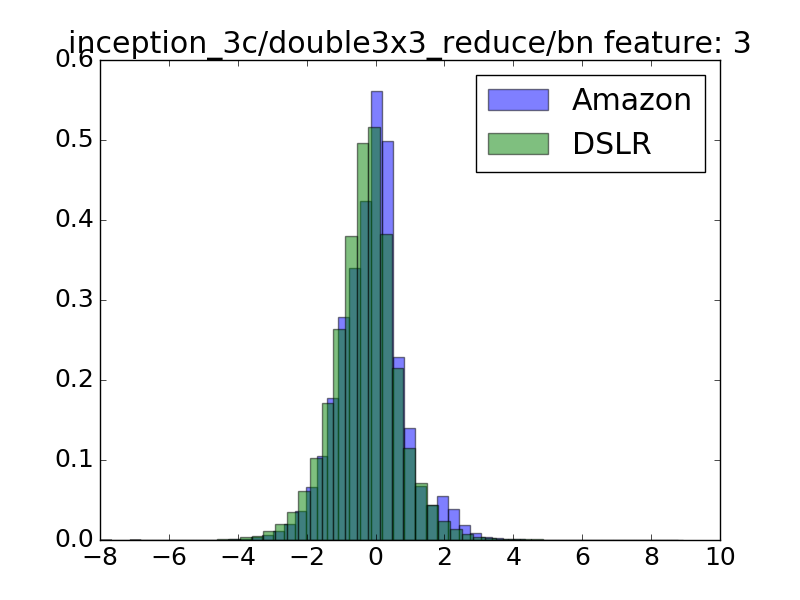}
  \caption{Distributions of randomly sampled source/target features from different layers of \DIALInception{} learned on the Amazon--DSLR task of the Office 31 dataset (best viewed on screen).}
  \label{fig:distr}
\end{figure*}

In this section we study the distributions of a set of randomly sampled features from different layers of \DIALInception{}, learned on the Amazon--DSLR task of the Office 31 dataset.
In Fig.~\ref{fig:distr} we compare the histograms of these features, computed on the whole source and target sets and taken \emph{after} the \DALs.
The plots clearly show the aligning effect of our \DALs, as most histograms are very closely matching.
It is also interesting to note how the alignment effect seems to be mostly independent of the particular shape the distributions might take.
\end{appendices}